%% file: paper.tex
\documentclass[10pt,twocolumn,letterpaper]{article}

\usepackage[utf8]{inputenc} %
\usepackage[T1]{fontenc}    %
\usepackage{url}            %
\usepackage{booktabs}       %
\usepackage{amsmath}
\usepackage{amsfonts}       %
\usepackage{nicefrac}       %
\usepackage{microtype}      %
\usepackage[pdftex]{graphicx}
\usepackage{placeins}
\usepackage{color}
\usepackage{enumitem}
\usepackage{cvpr}
\usepackage{times}
\usepackage{array}
\usepackage{bibunits}
\usepackage{chngcntr}
\usepackage{hyperref}
\def\cvprPaperID{4225} %

\newcommand{\appref}[1]{Appendix~\ref{#1}}

\title{Do Better ImageNet Models Transfer Better?}

\author{
  Simon Kornblith\thanks{Work done as a member of the Google AI Residency program
(\url{g.co/airesidency}).}, Jonathon Shlens, and Quoc V. Le\\
  Google Brain\\
  \texttt{\{skornblith,shlens,qvl\}@google.com}
}

\begin{document}

\cvprfinalcopy
\maketitle

\begin{abstract}

Transfer learning is a cornerstone of computer vision, yet little work has been done to evaluate the relationship between architecture and transfer. An implicit hypothesis in modern computer vision research is that models that perform better on ImageNet necessarily perform better on other vision tasks.
However, this hypothesis has never been systematically tested.
Here, we compare the performance of 16 classification networks on 12 image classification datasets. We find that, when networks are used as fixed feature extractors or fine-tuned, there is a strong correlation between ImageNet accuracy and transfer accuracy ($r = 0.99$ and $0.96$, respectively). In the former setting, we find that this relationship is very sensitive to the way in which networks are trained on ImageNet; many common forms of regularization slightly improve ImageNet accuracy but yield penultimate layer features that are much worse for transfer learning.
Additionally, we find that, on two small fine-grained image classification datasets, pretraining on ImageNet provides minimal benefits, indicating the learned features from ImageNet do not transfer well to fine-grained tasks.
Together, our results show that ImageNet architectures generalize well across datasets, but ImageNet features are less general than previously suggested.
\end{abstract}

\section{Introduction}
\begingroup

The last decade of computer vision research has pursued academic benchmarks as a measure of progress. No benchmark has been as hotly pursued as ImageNet \cite{deng2009imagenet,Russakovsky2015}. Network architectures measured against this dataset have fueled much progress in computer vision research across a broad array of problems, including transferring to new datasets \cite{donahue2014decaf,razavian2014cnn}, object detection \cite{huang2016}, image segmentation \cite{he2017mask,chen2018deeplab} and perceptual metrics of images \cite{johnson2016perceptual}. An implicit assumption behind this progress is that network architectures that perform better on ImageNet necessarily perform better on other vision tasks. Another assumption is that better network architectures learn better features that can be transferred across vision-based tasks.
Although previous studies have provided some evidence for these hypotheses (e.g. \cite{Chatfield14,simonyan2014very,huang2016,howard2017mobilenets,he2017mask}), they have never been systematically explored across network architectures.

\begin{figure}[t]
    \centering
    \includegraphics{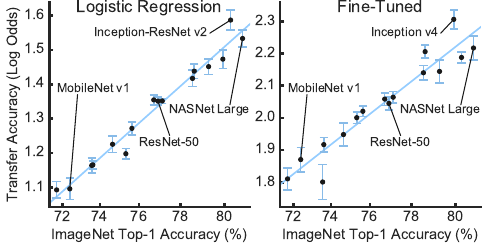}
    \caption{Transfer learning performance is highly correlated with ImageNet top-1 accuracy for fixed ImageNet features (left) and fine-tuning from ImageNet initialization (right). The 16 points in each plot represent transfer accuracy for 16 distinct CNN architectures, averaged across 12 datasets after logit transformation (see Section~\ref{methods}). Error bars measure variation in transfer accuracy across datasets. These plots are replicated in Figure \ref{scatterplots} (right).}
    \label{fig1}
\end{figure}

In the present work, we seek to test these hypotheses by investigating the transferability of both ImageNet features and ImageNet classification architectures. 
Specifically, we conduct a large-scale study of transfer learning across 16 modern convolutional neural networks for image classification on 12 image classification datasets in 3 different experimental settings: as fixed feature extractors \cite{donahue2014decaf,razavian2014cnn}, fine-tuned from ImageNet initialization \cite{agrawal2014,girshick2014rich,Chatfield14}, and trained from random initialization. Our main contributions are as follows:

\begin{itemize}
  \item Better ImageNet networks provide better penultimate layer features for transfer learning with linear classification ($r = 0.99$), and better performance when the entire network is fine-tuned ($r = 0.96$).
  \item Regularizers that improve ImageNet performance are highly detrimental to the performance of transfer learning based on penultimate layer features.
  \item Architectures transfer well across tasks even when weights do not. On two small fine-grained classification datasets, fine-tuning does not provide a substantial benefit over training from random initialization, but better ImageNet architectures nonetheless obtain higher accuracy.
\end{itemize}

\section{Related work}

ImageNet follows in a succession of progressively larger and more realistic benchmark datasets for computer vision. Each successive dataset was designed to address perceived issues with the size and content of previous datasets. Torralba and Efros \cite{torralba2011unbiased} showed that many early datasets were heavily biased, with classifiers trained to recognize or classify objects on those datasets possessing almost no ability to generalize to images from other datasets.

Early work using convolutional neural networks (CNNs) for transfer learning extracted fixed features from ImageNet-trained networks and used these features to train SVMs and logistic regression classifiers for new tasks \cite{donahue2014decaf,razavian2014cnn,Chatfield14}. These features could outperform hand-engineered features even for tasks very distinct from ImageNet classification \cite{donahue2014decaf,razavian2014cnn}. Following this work, several studies compared the performance of AlexNet-like CNNs of varying levels of computational complexity in a transfer learning setting with no fine-tuning. Chatfield et al. \cite{Chatfield14} found that, out of three networks, the two more computationally expensive networks performed better on PASCAL VOC. Similar work concluded that deeper networks produce higher accuracy across many transfer tasks, but wider networks produce lower accuracy \cite{razavian2016}. More recent evaluation efforts have investigated transfer from modern CNNs to medical image datasets \cite{mormont2018comparison}, and transfer of sentence embeddings to language tasks \cite{conneau2018senteval}.

A substantial body of existing research indicates that, in image tasks, fine-tuning typically achieves higher accuracy than classification based on fixed features, especially for larger datasets or datasets with a larger domain mismatch from the training set \cite{agrawal2014,Chatfield14,girshick2014rich,yosinski2014transferable,razavian2016,lin2015bilinear,HuhAE16,chu16,mormont2018comparison}. In object detection, ImageNet-pretrained networks are used as backbone models for Faster R-CNN and R-FCN detection systems \cite{ren2015faster,dai2016r}. Classifiers with higher ImageNet accuracy achieve higher overall object detection accuracy \cite{huang2016}, although variability across network architectures is small compared to variability from other object detection architecture choices. A parallel story likewise appears in image segmentation models \cite{chen2018deeplab}, although it has not been as systematically explored.
\endgroup

Several authors have investigated how properties of the original training dataset affect transfer accuracy. Work examining the performance of fixed image features drawn from networks trained on subsets of ImageNet have reached conflicting conclusions regarding the importance of the number of classes vs. number of images per class  \cite{HuhAE16,razavian2016}. Yosinski et al. \cite{yosinski2014transferable} showed that the first layer of AlexNet can be frozen when transferring between natural and manmade subsets of ImageNet without performance impairment, but freezing later layers produces a substantial drop in accuracy.
Other work has investigated transfer from extremely large image datasets to ImageNet, demonstrating that transfer learning can be useful even when the target dataset is large \cite{sun2017revisiting,mahajan2018exploring}. Finally, a recent work devised a strategy to transfer when labeled data from many different domains is available \cite{zamir2018taskonomy}.

\begin{table*}[htbp]
    \begin{center}
        \begin{tabular}{|l|l|r|r|r|r|l|}
        \hline
        Dataset  &  \multicolumn{1}{|l|}{Classes} & \multicolumn{1}{|l|}{Size (train/test)} & Accuracy metric\\
        \hline\hline
        Food-101 \cite{bossard2014food} & 101 & 75,750/25,250 & top-1\\
        CIFAR-10 \cite{krizhevsky2009learning} & 10 & 50,000/10,000 & top-1\\
        CIFAR-100 \cite{krizhevsky2009learning}& 100 & 50,000/10,000 & top-1\\
        Birdsnap \cite{berg2014birdsnap} & 500 & 47,386/2,443 & top-1\\
        SUN397 \cite{xiao2010sun} & 397 & 19,850/19,850 & top-1\\
        Stanford Cars \cite{krause2013collecting} & 196 & 8,144/8,041 & top-1\\
        FGVC Aircraft \cite{maji13fine-grained} & 100 & 6,667/3,333 & mean per-class\\
        PASCAL VOC 2007 Cls. \cite{everingham2010pascal} & 20 & 5,011/4,952 & 11-point mAP\\
        Describable Textures (DTD) \cite{cimpoi2014describing} & 47 & 3,760/1,880 & top-1\\
        Oxford-IIIT Pets \cite{parkhi2012cats} & 37 & 3,680/3,369 & mean per-class\\
        Caltech-101 \cite{fei2004learning} & 102 & 3,060/6,084 & mean per-class\\
        Oxford 102 Flowers \cite{nilsback2008automated} & 102 & 2,040/6,149 & mean per-class\\
        \hline
        \end{tabular}
    \end{center}
    \caption{Datasets examined in transfer learning}
    \label{table:dataset}
    \label{datasets}
\end{table*}

\section{Statistical methods}
\label{methods}

Much of the analysis in this work requires comparing accuracies across datasets of differing difficulty.  
 When fitting linear models to accuracy values across multiple datasets, we consider effects of model and dataset to be additive. In this context, using untransformed accuracy as a dependent variable is problematic: The meaning of a 1\% additive increase in accuracy is different if it is relative to a base accuracy of 50\% vs. 99\%. Thus, we consider the log odds, i.e., the accuracy after the logit transformation $\text{logit}(p) = \log(p/(1-p)) = \text{sigmoid}^{-1}(p)$. The logit transformation is the most commonly used transformation for analysis of proportion data, and an additive change $\Delta$ in logit-transformed accuracy has a simple interpretation as a multiplicative change $\exp \Delta$ in the odds of correct classification:
\begin{align*}
    \text{logit}\left(\frac{n_\text{correct}}{n_\text{correct}+n_\text{incorrect}}\right) + \Delta &= \log\left(\frac{n_\text{correct}}{n_\text{incorrect}}\right) + \Delta\\
    &= \log\left(\frac{n_\text{correct}}{n_\text{incorrect}} \exp \Delta\right)
\end{align*}
We plot all accuracy numbers on logit-scaled axes.

We computed error bars for model accuracy averaged across datasets, using the procedure from Morey \cite{morey2008} to remove variance due to inherent differences in dataset difficulty. Given logit-transformed accuracies $x_{md}$ of model $m \in \mathcal{M}$ on dataset $d \in \mathcal{D}$, we compute adjusted accuracies
 $\text{acc}(m, d) = x_\text{md} - \sum_{n \in \mathcal{M}} x_\text{nd}/|\mathcal{M}|$. For each model, we take the mean and standard error of the adjusted accuracy across datasets, and multiply the latter by a correction factor $\sqrt{|\mathcal{M}|/(|\mathcal{M}|-1)}$.

When examining the strength of the correlation between ImageNet accuracy and accuracy on transfer datasets, we report $r$ for the correlation between the logit-transformed ImageNet accuracy and the logit-transformed transfer accuracy averaged across datasets. We report the rank correlation (Spearman's $\rho$) in \appref{spearman_appendix}.

We tested for significant differences between pairs of networks on the same dataset using a permutation test or equivalent binomial test of the null hypothesis that the predictions of the two networks are equally likely to be correct, described further in  \appref{two_models_same_dataset}. We tested for significant differences between networks in average performance across datasets using a t-test.

\section{Results}

We examined 16 modern networks ranging in ImageNet (ILSVRC 2012 validation) top-1 accuracy from 71.6\% to 80.8\%. These networks encompassed widely used Inception architectures \cite{szegedy2015going,ioffe2015batch,szegedy2016rethinking,szegedy2017inception}; ResNets \cite{he2016deep,resnetv15torchblogpost,goyal2017accurate}; DenseNets \cite{huang2017densely}; MobileNets \cite{howard2017mobilenets,sandler2018}; and NASNets \cite{zoph2017learning}. For fair comparison, we retrained all models with scale parameters for batch normalization layers and without label smoothing, dropout, or auxiliary heads, rather than relying on pretrained models. \appref{model_info} provides training hyperparameters along with further details of each network, including the ImageNet top-1 accuracy, parameter count, dimension of the penultimate layer, input image size, and performance of retrained models. For all experiments, we rescaled images to the same image size as was used for ImageNet training.

We evaluated models on 12 image classification datasets ranging in training set size from 2,040 to 75,750 images (20 to 5,000 images per class; Table \ref{table:dataset}). These datasets covered a wide range of image classification tasks, including superordinate-level object classification (CIFAR-10 \cite{krizhevsky2009learning}, CIFAR-100 \cite{krizhevsky2009learning}, PASCAL VOC 2007 \cite{everingham2010pascal}, Caltech-101 \cite{fei2004learning}); fine-grained object classification (Food-101 \cite{bossard2014food}, Birdsnap \cite{berg2014birdsnap}, Stanford Cars \cite{krause2013collecting}, FGVC Aircraft \cite{maji13fine-grained}, Oxford-IIIT Pets \cite{parkhi2012cats}); texture classification (DTD \cite{cimpoi2014describing}); and scene classification (SUN397 \cite{xiao2010sun}).

\begin{figure*}[htbp]
    \centering
    \includegraphics[width=\linewidth]{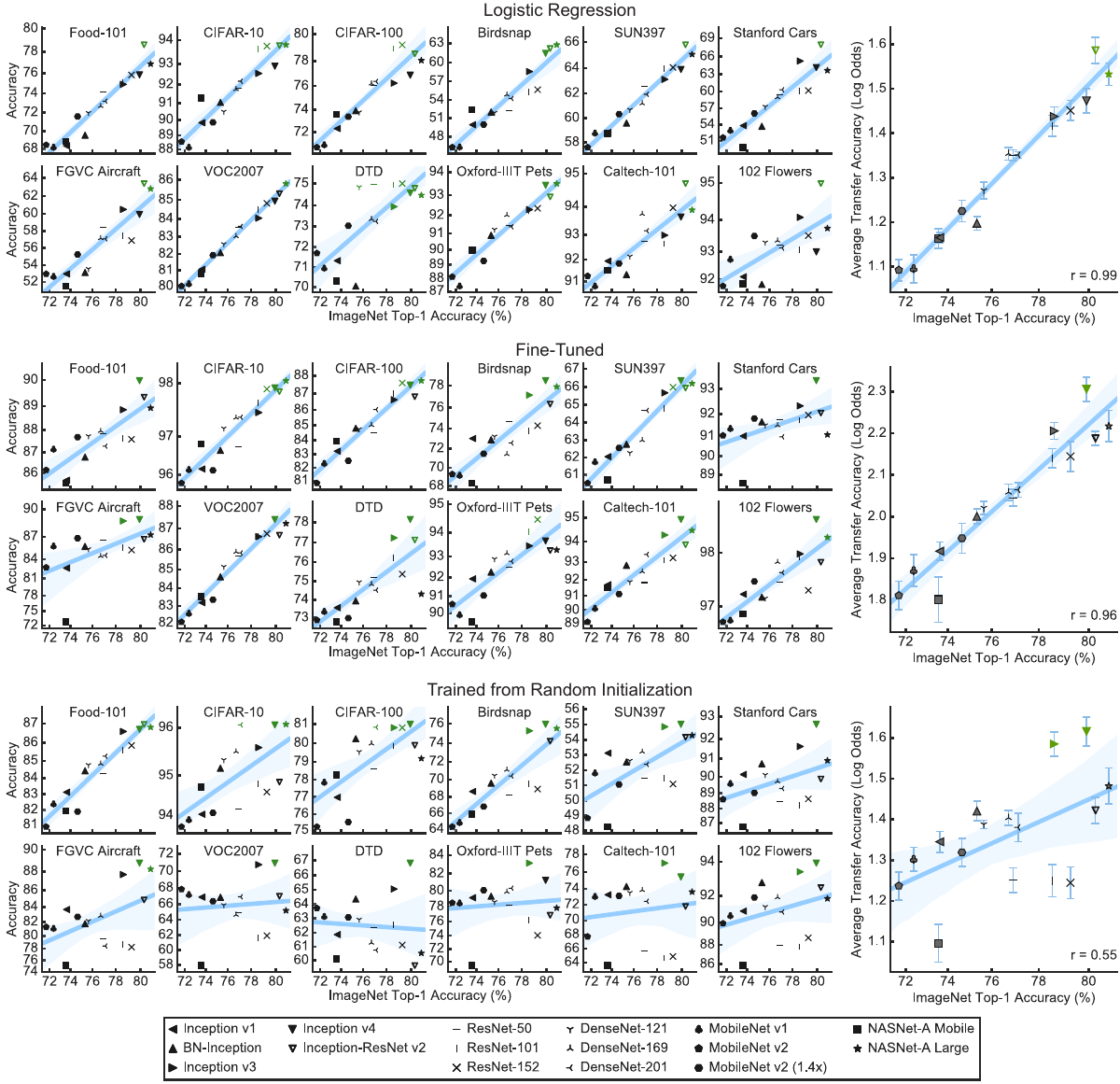}
    \caption{ImageNet accuracy is a strong predictor of transfer accuracy for logistic regression on penultimate layer features and fine-tuning. Each set of panels measures correlations between ImageNet accuracy and transfer accuracy across fixed ImageNet features (top), fine-tuned networks (middle) and networks trained from scratch (bottom). Left: Relationship between classification accuracy on transfer datasets (y-axis) and ImageNet top-1 accuracy (x-axis) in different training settings. Axes are logit-scaled (see text). The regression line and a 95\% bootstrap confidence interval are plotted in blue. Right: Average log odds of correct classification across datasets. Error bars are standard error. Points corresponding to models not significantly different from the best model ($p > 0.05$) are colored green.}
    \label{scatterplots}
\end{figure*}

Figure \ref{scatterplots} presents correlations between the top-1 accuracy on ImageNet vs. the performance of the same model architecture on new image tasks. We measure transfer learning performance in three settings: (1) training a logistic regression classifier on the {\it fixed} feature representation from the penultimate layer of the ImageNet-pretrained network, (2) fine-tuning the ImageNet-pretrained network, and (3) training the same CNN architecture from scratch on the new image task.

\subsection{ImageNet accuracy predicts performance of logistic regression on fixed features, but regularization settings matter}

We first examined the performance of different networks when used as {\it fixed} feature extractors by training an $L_2$-regularized logistic regression classifier on penultimate layer activations using L-BFGS \cite{liu1989limited} without data augmentation.\footnote{We also repeated these experiments with support vector machine classifiers in place of logistic regression, and when using data augmentation for logistic regression; see \appref{svm}. Findings did not change.} As shown in Figure~\ref{scatterplots} (top), ImageNet top-1 accuracy was highly correlated with accuracy on transfer tasks ($r = 0.99$). Inception-ResNet v2 and NASNet Large, the top two models in terms of ImageNet accuracy, were statistically tied for first place.

\begin{figure}[htbp]
    \centering
    \includegraphics{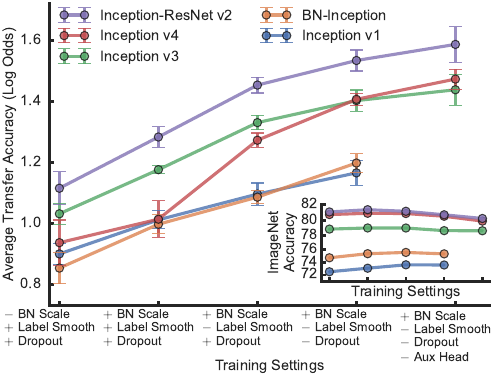}
    \caption{ImageNet training settings have a large effect upon performance of logistic regression classifiers trained on penultimate layer features. In the main plot, each point represents the logit-transformed transfer accuracy averaged across the 12 datasets, measured using logistic regression on penultimate layer features from a specific model trained with the training configuration labeled at the bottom. "$+$" indicates that a setting was enabled, whereas "$-$" indicates that a setting was disabled. The leftmost, most heavily regularized configuration is typically used for Inception models \cite{szegedy2016rethinking}; the rightmost is typically used for ResNets and DenseNets. The inset plot shows ImageNet top-1 accuracy for the same training configurations. See also \appref{checkpoints_appendix}. Best viewed in color.}
    \label{model_variants}
\end{figure}

Critically, results in Figure~\ref{scatterplots} were obtained with models that were all trained on ImageNet with the same training settings. In experiments conducted with publicly available checkpoints, we were surprised to find that ResNets and DenseNets consistently achieved higher accuracy than other models, and the correlation between ImageNet accuracy and transfer accuracy with fixed features was low and not statistically significant (\appref{public_logreg}). Further investigation revealed that the poor correlation arose from differences in regularization used for these public checkpoints.

\begin{figure}
    \centering
    \includegraphics{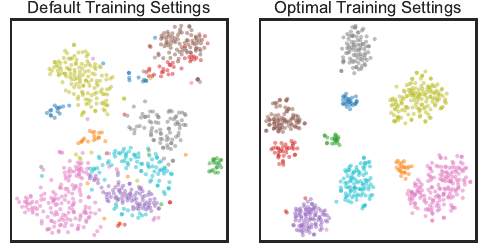}
    \caption{The default Inception training settings produce a suboptimal feature space. Low dimensional embeddings of Oxford 102 Flowers using t-SNE \cite{maaten2008visualizing} on features from the penultimate layer of Inception v4, for 10 classes from the test set. Best viewed in color.}
    \label{tsne}
\end{figure}

Figure~\ref{model_variants} shows the transfer learning performance of Inception models with different training settings. We identify 4 choices made in the Inception training procedure and subsequently adopted by several other models that are detrimental to transfer accuracy: (1) The absence of scale parameter $(\gamma)$ for batch normalization layers; the use of (2) label smoothing \cite{szegedy2016rethinking} and (3) dropout \cite{srivastava2014dropout}; and (4) the presence of an auxiliary classifier head \cite{szegedy2015going}. These settings had a small ($<1\%$) impact on the overall ImageNet top-1 accuracy of each model (Figure \ref{model_variants}, inset). However, in terms of average transfer accuracy, the difference between the default and optimal training settings was approximately equal to the difference between the worst and best ImageNet models trained with optimal settings. This difference was visible not only in transfer accuracy, but also in t-SNE embeddings of the features (Figure~\ref{tsne}). Differences in transfer accuracy between settings were apparent earlier in training than differences in ImageNet accuracy, and were consistent across datasets (\appref{checkpoints_appendix}).

Label smoothing and dropout are regularizers in the traditional sense: They are intended to improve generalization accuracy at the expense of training accuracy. Although auxiliary classifier heads were initially proposed to alleviate issues related to vanishing gradients \cite{lee2015deeply,szegedy2015going}, Szegedy et al. \cite{szegedy2016rethinking} instead suggest that they also act as regularizers. The improvement in transfer performance when incorporating batch normalization scale parameters may relate to changes in effective learning rates \cite{laarhoven2017,zhang2018}.

\subsection{ImageNet accuracy predicts fine-tuning performance}

We also examined performance when fine-tuning ImageNet networks (Figure~\ref{scatterplots}, middle). We initialized each network from the ImageNet weights and fine-tuned for 20,000 steps with Nesterov momentum and a cosine decay learning rate schedule at a batch size of 256. We performed grid search to select the optimal learning rate and weight decay based on a validation set (for details, see \appref{finetune_appendix}). Again, we found that ImageNet top-1 accuracy was highly correlated with transfer accuracy ($r = 0.96$).

\begin{figure}[t]
    \centering
    \includegraphics{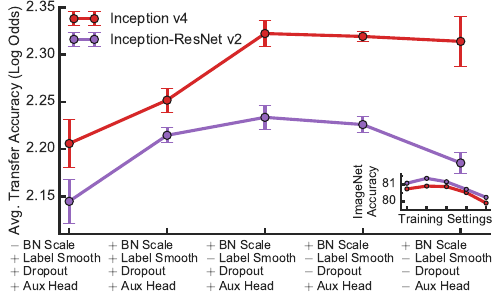}
    \caption{ImageNet training settings have only a minor impact on fine-tuning performance. Each point represents transfer accuracy for a model pretrained and fine-tuned with the same training configuration, labeled at the bottom. Axes follow Figure \ref{model_variants}. See \appref{finetune_reg_appendix} for performance of models pretrained with regularization but fine-tuned without regularization.}
    \label{ft_settings}
\end{figure}

Compared with the logistic regression setting, regularization and training settings had smaller effects upon the performance of fine-tuned models. Figure~\ref{ft_settings} shows average transfer accuracy for Inception v4 and Inception-ResNet v2 models with different regularization settings. As in the logistic regression setting, introducing a batch normalization scale parameter and disabling label smoothing improved performance. In contrast to the logistic regression setting, dropout and the auxiliary head sometimes improved performance, but only if used during fine-tuning. We discuss these results further in \appref{finetune_reg_appendix}.

\begin{figure*}[htbp]
    \centering
    \includegraphics{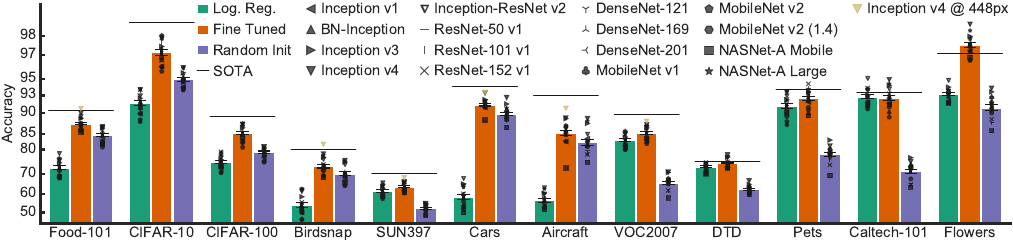}
    \caption{Performance comparison of logistic regression, fine-tuning, and training from random initialization. Bars reflect accuracy across models (excluding VGG) for logistic regression, fine-tuning, and training from random initialization. Error bars are standard error. Points represent individual models. Lines represent previous state-of-the-art. Best viewed in color.}
    \label{dataset_bars}
\end{figure*}

Overall, fine-tuning yielded better performance than classifiers trained on fixed ImageNet features, but the gain differed by dataset. Fine-tuning improved performance over logistic regression in 179 out of 192 dataset and model combinations (Figure \ref{dataset_bars}; see also \appref{comparison_scatters_appendix}). When averaged across the tested architectures, fine-tuning yielded significantly better results on all datasets except Caltech-101 (all $p < 0.01$, Wilcoxon signed rank test; Figure \ref{dataset_bars}). The improvement was generally larger for larger datasets. However, fine-tuning provided substantial gains on the smallest dataset, 102 Flowers, with 102 classes and 2,040 training examples.

\subsection{ImageNet accuracy predicts performance of networks trained from random initialization}

One confound of the previous results is that it is not clear whether ImageNet accuracy for transfer learning is due to the weights derived from the ImageNet training or the architecture itself. To remove the confound, we next examined architectures trained from random initialization, using a similar training setup as for fine-tuning (see \appref{randominit_appendix}). In this setting, the correlation between ImageNet top-1 accuracy and accuracy on the new tasks was more variable than in the transfer learning settings, but there was a tendency toward higher performance for models that achieved higher accuracy on ImageNet ($r = 0.55$; Figure~\ref{scatterplots}, bottom).
 
Examining these results further, we found that a single correlation averages over a large amount of variability. For the 7 datasets with <10,000 examples, the correlation was low and did not reach statistically significance ($r = 0.29$; see also  \appref{second_order_appendix}). However, for the larger datasets, the correlation between ImageNet top-1 accuracy and transfer learning performance was markedly stronger ($r = 0.86$). Inception v3 and v4 were among the top-performing models across all dataset sizes.

\subsection{Benefits of better models are comparable to specialized methods for transfer learning}

 Given the strong correlation between ImageNet accuracy and transfer accuracy, we next sought to compare simple approaches to transfer learning with better ImageNet models with baselines from the literature. We achieve state-of-the-art performance on half of the 12 datasets if we evaluate using the same image sizes as the baseline methods (Figure \ref{dataset_bars}; see full results in \appref{best_models_appendix}). Our results suggest that the ImageNet performance of the pretrained model is a critical factor in transfer performance. %
 
 Several papers have proposed methods to make better use of CNN features and thus improve the efficacy of transfer learning \cite{lin2015bilinear,cimpoi2015deep,lin2016visualizing,gao2016compact,yao2016coarse,song2017locally,cui2017kernel,li2017dynamic,peng2018object}. On the datasets we examine, we outperform all such methods simply by fine-tuning state-of-the-art CNNs (\appref{best_models_appendix}). Moreover, in some cases a better CNN can make up for dataset deficiencies: By fine-tuning ImageNet-pretrained Inception v4, we outperform the best reported single-model results for networks pretrained on the Places dataset \cite{herranz2016scene,zhou2017places}, which more closely matches the domain of SUN397.
 
 It is likely that improvements obtained with better models, specialized transfer learning methods, and pretraining datasets with greater domain match are complementary. Combining these approaches could lead to even better performance. Nonetheless, it is surprising that simply using a better model can yield gains comparable to specialized techniques.

\subsection{ImageNet pretraining does not necessarily improve accuracy on fine-grained tasks}

\begin{figure}[t]
    \centering
    \includegraphics{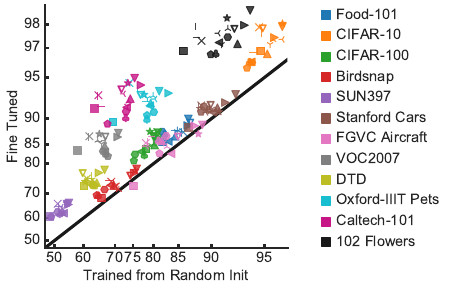}
    \caption{For some datasets and networks, the gap between fine-tuning and training from random initialization is small. Each point represents a dataset/model combination. Axes are logit-scaled. See Figure~\ref{dataset_bars} for network legend and \appref{comparison_scatters_appendix} for scatter plots of other settings. Best viewed in color.}
    \label{ft_scratch}
\end{figure}

Fine-tuning was more accurate than training from random initialization for 189 out of 192 dataset/model combinations, but on Stanford Cars and FGVC Aircraft, the improvement was unexpectedly small (Figures \ref{dataset_bars} and \ref{ft_scratch}). In both settings, Inception v4 was the best model we tested on these datasets. When trained at the default image size of $299 \times 299$, it achieved 92.7\% on Stanford Cars when trained from scratch on vs. 93.3\% when fine-tuned, and 88.8\% on FGVC Aircraft when trained from scratch vs. 89.0\% when fine-tuned.

\begin{figure*}[htbp]
    \centering
    \includegraphics[width=\linewidth]{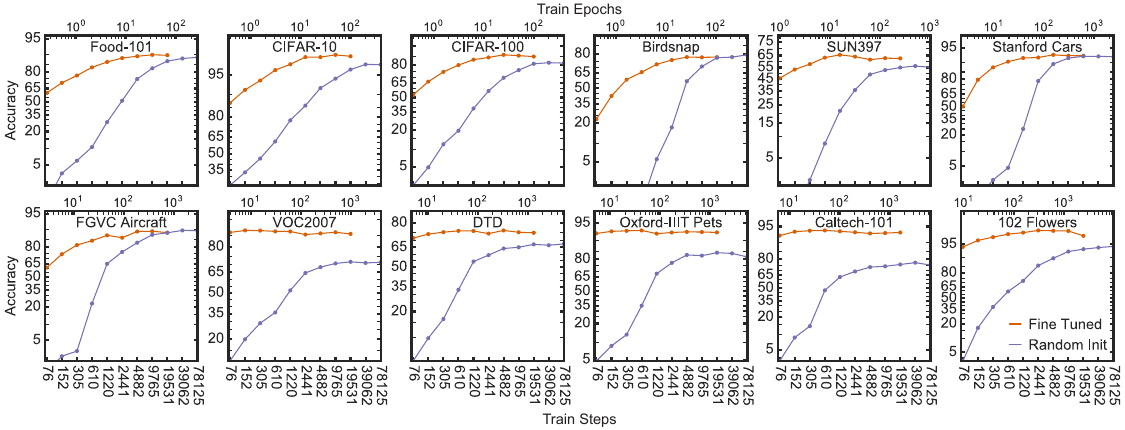}
    \caption{Networks pretrained on ImageNet converge faster, even when final accuracy is the same as training from random initialization. Each point represents an independent Inception v4 model trained with optimized hyperparameters. For fine-tuning, we initialize with the public TensorFlow Inception v4 checkpoint. Axes are logit-scaled.}
    \label{train_time}
\end{figure*}

ImageNet pretraining thus appears to have only marginal accuracy benefits for fine-grained classification tasks where labels are not well-represented in ImageNet. At 100+ classes and <10,000 examples, Stanford Cars and FGVC Aircraft are much smaller than most datasets used to train CNNs \cite{jeffdean}. In fact, the ImageNet training set contains more car images than Stanford Cars (12,455 vs. 8,144). However, ImageNet contains only 10 high-level car classes (e.g., sports car), whereas Stanford Cars contains 196 car classes by make, model, and year.
Four other datasets (Oxford 102 Flowers, Oxford-IIIT Pets, Birdsnap, and Food-101) require similarly fine-grained classification, but the classes contained in the latter three datasets are much better-represented in ImageNet. Most of the cat and dog breeds present in Oxford-IIIT Pets correspond directly to ImageNet classes, and ImageNet contains 59 classes of birds and around 45 classes of fruits, vegetables, and prepared dishes.%

\subsection{ImageNet pretraining accelerates convergence}
\label{train_time_section}

Given that fine-tuning and training from random initialization achieved similar performance on Stanford Cars and FGVC Aircraft, we next asked whether fine-tuning still posed an advantage in terms of training time. In Figure \ref{train_time}, we examine performance of Inception v4 when fine-tuning or training from random initialization for different numbers of steps. Even when fine-tuning and training from scratch achieved similar final accuracy, we could fine-tune the model to this level of accuracy in an order of magnitude fewer steps. To quantify this acceleration, we computed the number of epochs and steps required to reach 90\% of the maximum odds of correct classification achieved at any number of steps, and computed the geometric mean across datasets. Fine-tuning reached this threshold level of accuracy in an average of 26 epochs/1151 steps (inter-quartile ranges 267-4882 steps, 12-58 epochs), whereas training from scratch required 444 epochs/19531 steps (inter-quartile ranges 9765-39062 steps, 208-873 epochs) corresponding to a 17-fold speedup on average.

\begin{figure*}[htbp]
    \centering
    \includegraphics[width=\linewidth]{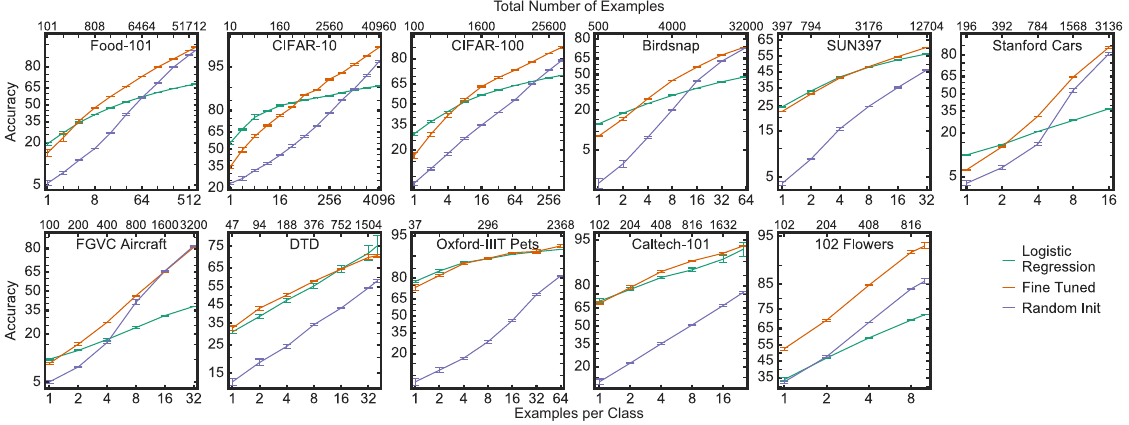}
    \caption{Pretraining on ImageNet improves performance on fine-grained tasks with small amounts of data, but the gap narrows quickly as dataset size increases. Performance of transfer learning with the public Inception v4 model at different dataset sizes. Error bars reflect standard error over 3 subsets. Note that the maximum dataset size shown is not the full dataset. Best viewed in color.}
    \label{acc_vs_examples}
\end{figure*}

\subsection{Accuracy benefits of ImageNet pretraining fade quickly with dataset size}
\label{dataset_size_section}

Although all datasets benefit substantially from ImageNet pretraining when few examples are available for transfer, for many datasets, these benefits fade quickly when more examples are available. In Figure \ref{acc_vs_examples}, we show the behavior of logistic regression, fine-tuning, and training from random initialization in the regime of limited data, i.e., for dataset subsets consisting of different numbers of examples per class. When data is sparse (47-800 total examples), logistic regression is a strong baseline, achieving accuracy comparable to or better than fine-tuning. At larger dataset sizes, fine-tuning achieves higher performance than logistic regression, and, for fine-grained classification datasets, the performance of training from random initialization begins to approach results of pre-trained models. On FGVC Aircraft, training from random initialization achieved parity with fine-tuning at only 1600 total examples (16 examples per class).

\section{Discussion}

Has the computer vision community overfit to ImageNet as a dataset? In a broad sense, our results suggest the answer is no: We find that there is a strong correlation between ImageNet top-1 accuracy and transfer accuracy, suggesting that better ImageNet architectures are capable of learning better, transferable representations. But we also find that a number of widely-used regularizers that improve ImageNet performance do not produce better representations. These regularizers are harmful to the penultimate layer feature space, and have mixed effects when networks are fine-tuned.

More generally, our results reveal clear limits to transferring features, even among natural image datasets. ImageNet pretraining accelerates convergence and improves performance on many datasets, but its value diminishes with greater training time, more training data, and greater divergence from ImageNet labels. For some fine-grained classification datasets, a few thousand labeled examples, or a few dozen per class, are all that are needed to make training from scratch perform competitively with fine-tuning. Surprisingly, however, the value of architecture persists.

The last decade of computer vision research has demonstrated the superiority of image features learned from data over generic, hand-crafted features. Before the rise of convolutional neural networks, most approaches to image understanding relied on hand-engineered feature descriptors \cite{lowe1999object,dalal2005histograms,bay2008speeded}.
Krizhevsky et al. \cite{krizhevsky2012imagenet} showed that, given the training data provided by ImageNet \cite{deng2009imagenet}, features learned by convolutional neural networks could substantially outperform these hand-engineered features. Soon after, it became clear that intermediate representations learned from ImageNet also provided substantial gains over hand-engineered features when transferred to other tasks \cite{donahue2014decaf,razavian2014cnn}.

Is the general enterprise of learning widely-useful features doomed to suffer the same fate as feature engineering? Given differences between datasets \cite{torralba2011unbiased}, it is not entirely surprising that features learned on one dataset benefit from some amount of adaptation when applied to another. However, given the history of attempts to build general natural-image feature descriptors, it is surprising that common transfer learning approaches cannot always profitably adapt features learned from a large natural-image to a much smaller natural-image dataset.

ImageNet weights provide a starting point for features on a new classification task, but perhaps what is needed is a way to learn adaptable features. This problem is closely related to few-shot learning \cite{lake2015human,vinyals2016matching,ravi2016optimization,snell2017prototypical,finn2017model,snell2017prototypical,mishra2017meta}, but these methods are typically evaluated with training and test classes from the same distribution. Common few-shot learning methods do not seem to outperform classifiers trained on fixed features when domain shift is present \cite{chen2018a}, but it may be possible to obtain better results with specialized methods \cite{dong2018domain} or by combining few-shot learning methods with fine-tuning \cite{scott2018adapted}. It thus remains to be seen whether methods can be developed or repurposed to adapt visual representations learned from ImageNet to provide larger benefits across natural image tasks.

\section*{Acknowledgements}
We thank George Dahl, Boyang Deng, Sara Hooker, Pieter-jan Kindermans, Rafael Müller, Jiquan Ngiam, Ruoming Pang, Daiyi Peng, Kevin Swersky, Vishy Tirumalashetty, Vijay Vasudevan, and Emily Xue for comments on the experiments and manuscript, and Aliza Elkin and members of the Google Brain team for support and ideas.

{\small
\bibliographystyle{ieee}
\bibliography{paper.bib}
}

\newpage
\appendix
\counterwithin{figure}{section}
\counterwithin{table}{section}
\include{appendix}

\end{document}

%% file: appendix.tex
\onecolumn
\begin{center}
  {\Large \bf Supplementary Material for ``Do Better ImageNet Models Transfer Better?" \par}
  \vspace*{24pt}
  {
  \large
  \lineskip .5em
  \begin{tabular}[t]{c}
      Simon Kornblith, Jonathon Shlens, and Quoc V. Le\\
      Google Brain\\
      \texttt{\{skornblith,shlens,qvl\}@google.com}
  \end{tabular}
  \par
  }
  \vskip .5em
  \vspace*{12pt}
\end{center}

\section{Supplementary experimental procedures}
\setcounter{page}{1}

\subsection{Statistical methods}

\subsubsection{Comparison of two models on the same dataset}
\label{two_models_same_dataset}
To test for superiority of one model over another on a given dataset, we constructed permutations where, for each example, we randomly exchanged the predictions of the two networks. For each permutation, we computed the difference in accuracy between the two networks. (For VOC2007, we considered the accuracy of predictions across labels.) We computed a p-value as the proportion of permutations where the difference is at least as extreme as the observed difference in accuracy. For top-1 accuracy, this procedure is equivalent to a binomial test sometimes called the "exact McNemar test," and a p-value can be computed exactly. For mean per-class accuracy, we approximated a p-value based on 10,000 permutations. These tests assess whether one trained model performs better than another on data drawn from the test set distribution. However, they are tests between trained models, rather than tests between architectures, since we do not measure variability arising from training networks from different random initializations or from different orderings of the training data.

\subsubsection{Measures of correlation}
\label{spearman_appendix}

\begin{table}[htbp]
    \begin{minipage}{\linewidth}
        \centering
        \small
        \begin{tabular}{|>{\raggedright}p{3cm}|r|r|r|r|}
        \hline
        \multicolumn{1}{|c|}{Setting} & \multicolumn{1}{c|}{$r^2$} & \multicolumn{1}{c|}{$r$} &  \multicolumn{1}{c|}{$\rho$} &  \multicolumn{1}{c|}{p-value}\\
        \hline\hline
        Logistic regression & 0.97 & 0.99 & 0.99 & $<10^{-11}$\\
        Fine-tuned & 0.91 & 0.96 & 0.97 & $<10^{-8}$\\
        Trained from scratch & 0.30 & 0.55 & 0.59 & 0.03\\
        Logistic regression (public checkpoints) & 0.14 & 0.37 & 0.48 & 0.16\\
        \hline
        \end{tabular}
    \end{minipage}
    \caption{Correlations between ImageNet accuracy and average transfer accuracy (Pearson $r$ and $r^2$ and Spearman's $\rho$), as well as p-values for the null hypothesis that $r = 0$.}
    \label{correlations_table}
\end{table}

Table \ref{correlations_table} shows the Pearson correlation (as $r^2$ and $r$) as well as the Spearman rank correlation ($\rho$) in each of the three transfer settings we examine. We believe that Pearson correlation is the more appropriate measure, given that it is less dependent on the specific CNNs chosen for the study and the effects are approximately linear, but our results are similar in either case.

\subsection{Datasets}

All datasets had a median image size on the shortest side of at least 331 pixels (the highest ImageNet-native input image size out of all networks tested), except Caltech-101, for which the median size is 225 on the shortest side and 300 on the longer side, and CIFAR-10 and CIFAR-100, which consist of $32\times 32$ pixel images.

For datasets with a provided validation set (FGVC Aircraft, VOC2007, DTD, and 102 Flowers), we used this validation set to select hyperparameters. For other datasets, we constructed a validation set by subsetting the original training set. For the DTD and SUN397 datasets, which provide multiple train/test splits, we used only the first provided split. For the Caltech-101 dataset, which specifies no train/test split, we trained on 30 images per class and tested on the remainder, as in previous works \cite{donahue2014decaf,simonyan2014very,zeiler2014visualizing,Chatfield14}. With the exception of dataset subset results (Figure \ref{acc_vs_examples}), all results indicate the performance of models retrained on the combined training and validation set.

\subsection{Networks and ImageNet training procedure}

\label{model_info}

\begin{table}[h]
    \begin{minipage}{\linewidth}
        \centering
        \small
        \begin{tabular}{|l|r|r|r|r|r|r|}
        \hline
        \multicolumn{4}{|c|}{} & \multicolumn{3}{c|}{ImageNet Top-1 Accuracy}\\
        \multicolumn{1}{|c|}{Model} &  \multicolumn{1}{c|}{Parameters\footnote{Excludes logits layer.}} & \multicolumn{1}{c|}{Features} & \multicolumn{1}{c|}{Image Size} & \multicolumn{1}{c|}{Paper} & \multicolumn{1}{c|}{Public Checkpoint\footnote{Performance of checkpoint from TF-Slim repository (\texttt{https://github.com/tensorflow/models/tree/master/research/slim}), or, for DenseNets, from Keras applications (\texttt{https://keras.io/applications/}).}} & \multicolumn{1}{c|}{Retrained}\\
        \hline\hline
        Inception v1\footnote{We used Inception model code from the TF-Slim repository, which uses batch normalization layers for Inception v1. Additionally, the models in this repository contain minor modifications compared to the models described in the original papers. We cite the performance number for BN-GoogLeNet from Szegedy et al. \cite{szegedy2016rethinking}.} \cite{szegedy2015going} & 5.6M & 1024 & 224 & 73.2 & 69.8 & 73.6\\
        BN-Inception\footnote{This model is called "Inception v2" in TF-Slim model repository, but matches the model described in Ioffe and Szegedy \cite{ioffe2015batch}, rather than the model that Szegedy et al. \cite{szegedy2016rethinking} call "Inception v2."} \cite{ioffe2015batch} & 10.2M & 1024 & 224 & 74.8 & 74.0 & 75.3\\
        Inception v3 \cite{szegedy2016rethinking} & 21.8M & 2048 & 299 & 78.8 & 78.0 & 78.6\\
        Inception v4 \cite{szegedy2017inception} & 41.1M & 1536 & 299 & 80.0 & 80.2 & 79.9\\
        Inception-ResNet v2 \cite{szegedy2017inception} & 54.3M & 1536 & 299 & 80.1 & 80.4 & 80.3\\
        ResNet-50 v1\footnote{The ResNets we train incorporate two common modifications to the original ResNet v1 model: Stride-2 downsampling on the $3\times 3$ convolution instead of the first $1\times 1$ convolution in the block \cite{resnetv15torchblogpost,goyal2017accurate} and initialization of the batch normalization $\gamma$ to 0 in the last batch normalization layer of each block \cite{goyal2017accurate}. We report the numbers from Goyal et al. \cite{goyal2017accurate} as the original accuracy. No public TensorFlow checkpoints are available for these models, so, for public checkpoint results, we use the TF-Slim ResNet v1 checkpoints, which were converted from the original He et al. \cite{he2016deep} model.} \cite{he2016deep,resnetv15torchblogpost,goyal2017accurate} & 23.5M & 2048 & 224 & 76.4 & 75.2 & 76.9\\
        ResNet-101 v1 \cite{he2016deep,resnetv15torchblogpost,goyal2017accurate} & 42.5M & 2048 & 224 & 77.9 & 76.4 & 78.6\\
        ResNet-152 v1 \cite{he2016deep,resnetv15torchblogpost,goyal2017accurate} & 58.1M & 2048 & 224 & N/A & 76.8 & 79.3\\
        DenseNet-121 \cite{huang2017densely} & 7.0M & 1024 & 224 & 75.0 & 74.8 & 75.6\\
        DenseNet-169 \cite{huang2017densely} & 12.5M & 1664 & 224 & 76.2 & 76.2 & 76.7\\
        DenseNet-201 \cite{huang2017densely} & 18.1M & 1920 & 224 & 77.4 & 77.3 & 77.1\\
        MobileNet v1 \cite{howard2017mobilenets} & 3.2M & 1024 & 224 & 70.6 & 70.7 & 72.4\\
        MobileNet v2 \cite{sandler2018} & 2.2M & 1280 & 224 & 72.0 & 71.8 & 71.6\\
        MobileNet v2 (1.4) \cite{sandler2018} & 4.3M & 1792 & 224 & 74.7 & 75.0 & 74.7\\
        NASNet-A Mobile \cite{zoph2017learning} & 4.2M & 1056 & 224 & 74.0 & 74.0 & 73.6\\
        NASNet-A Large \cite{zoph2017learning} & 84.7M & 4032 & 331 & 82.7 & 82.7 & 80.8\\
        \hline
        \end{tabular}
    \end{minipage}
    \caption{ImageNet classification networks}
    \label{networks}
\end{table}

Table \ref{networks} lists the parameter count, penultimate layer feature dimension, and input image size for each network examined. Unless otherwise stated, our results were obtained with networks we trained, rather than publicly available checkpoints. We trained all networks with a batch size of 4096 using Nesterov momentum of 0.9 and weight decay of $8 \times 10^{-5}$, taking an exponential moving average of the weights with a decay factor of 0.9999. We performed linear warmup to a learning rate of 1.6 over the first 10 epochs, and then continuously decayed the learning rate by a factor of 0.975 per epoch. We used the preprocessing and data augmentation from \cite{inception_preprocessing}. To determine how long to train each network, we trained a separate model for up to 300 epochs with approximately 50,000 ImageNet training images held out as a validation set, and then trained a model on the full ImageNet training set for the number of steps that yielded the highest performance. Except in experiments explicitly studying the effects of these choices, for all networks, we used scale parameters for batch normalization layers, and did not use label smoothing, dropout, or an auxiliary head. For NASNet-A Large, we additionally disabled drop path regularization.

When training on ImageNet, we did not optimize hyperparameters for each network individually because we were able to achieve ImageNet top-1 performance comparable to publicly available checkpoints without doing so. (When fine-tuning and training from random initialization, we found that hyperparameters were more important and performed extensive tuning; see below.) For all networks except NASNet-A Large, our retrained models achieved accuracy no more than 0.5\% lower than the original reported results and public checkpoint, and sometimes substantially higher (Table~\ref{networks}). Given that we disabled the regularizers used in the original model, we expected a larger performance drop. Our experiments indicate that these regularizers further improve accuracy, but are evidently not necessary to achieve performance close to the published results.

For NASNet-A Large, there was a substantial gap between the performance of the published model and our retrained model (82.7\% vs. 80.8\%). As a sanity check, we enabled label smoothing, dropout, the auxiliary head, and drop path, and retrained NASNet-A Large with the same hyperparameters described above. This regularized model achieved 82.5\% accuracy, suggesting that most of the loss in accuracy in our setup is due to disabling regularization. For other models, we could further improve ImageNet top-1 accuracy over published results by applying regularizers: A retrained Inception-ResNet v2 model with label smoothing, dropout, and the auxiliary head enabled achieved 81.4\% top-1 accuracy, 1.1\% better than the unregularized model and 1.3\% better than the published result \cite{szegedy2017inception}.
However, because these regularizers clearly hurt results in the logistic regression setting, and because our goal was to compare all models and settings fairly, we report results for models trained and fine-tuned without regularization unless otherwise specified.

\subsection{Logistic regression}
\label{lr_appendix}

For each dataset, we extracted features from the penultimate layer of the network. We trained a multinomial logistic regression classifier using L-BFGS, with an L2 regularization parameter applied to the sum of the per-example losses, selected from a range of 45 logarithmically spaced values from $10^{-6}$ to $10^{5}$ on the validation set. Since the optimization problem is convex, we used the solution at the previous point along the regularization path as a warm start for the next point, which greatly accelerated the search. For these experiments, we did not perform data augmentation or scale aggregation, and we used the entire image, rather than cropping the central 87.5\% as is common for testing on ImageNet.

\subsection{Fine-tuning}
\label{finetune_appendix}

For fine-tuning experiments in Figure~\ref{scatterplots}, we initialized networks with ImageNet-pretrained weights and trained for 20,000 steps at a batch size of 256 using Nesterov momentum with a momentum parameter of 0.9. We selected the optimal learning rate and weight decay on the validation set by grid search. Our early experiments indicated that the optimal weight decay at a given learning rate varied inversely with the learning rate, as has been recently reported \cite{loschilov}. Thus, our grid consisted of 7 logarithmically spaced learning rates between 0.0001 and 0.1 and 7 logarithmically spaced weight decay to learning rate ratios between $10^{-6}$ and $10^{-3}$, as well as no weight decay. We found it useful to decrease the batch normalization momentum parameter from its ImageNet value to $\text{max}(1 - 10/s, 0.9)$ where $s$ is the number of steps per epoch. We found that the maximum performance on the validation set at any step during training was very similar to the maximum performance at the last step, presumably because we searched over learning rate, so we did not perform early stopping. On the validation set, we evaluated on both uncropped images and images cropped to the central 87.5\% and picked the approach that gave higher accuracy for evaluation on the test set. Cropped images typically yielded better performance, except on CIFAR-10 and CIFAR-100, where results differed by model.

When examining the effect of dataset size (Section \ref{dataset_size_section}), we fine-tuned for at least 1000 steps or 100 epochs (following guidance from our analysis of training time in Section \ref{train_time_section}) at a batch size of 64, with the learning rate range scaled down by a factor of 4. Otherwise, we used the same settings as above. Because we chose hyperparameters based on a large validation set, the results may not reflect what can be accomplished in practice when training on datasets of this size \cite{46794}. In Sections \ref{dataset_size_section} and \ref{train_time_section}, we fine-tuned models from the publicly available Inception v4 checkpoint rather than using the model trained as above.

\subsection{Training from random initialization}
\label{randominit_appendix}

We used a similar training protocol for training from random initialization as for fine-tuning, i.e., we trained for 20,000 steps at a batch size of 256 using Nesterov momentum with a momentum parameter of 0.9. Training from random initialization generally achieved optimal performance at higher learning rates and with greater weight decay, so we adjusted the learning rate range to span from 0.001 to 1.0 and the weight decay to learning rate ratio range to span from $10^{-5}$ to $10^{-2}$.

When examining the effect of dataset size (Section \ref{dataset_size_section}), we trained from random initialization for at least 78,125 steps or 200 epochs at a batch size of 16, with the learning rate range scaled down by a factor of 16. We chose these parameters because investigation of effects of training time (Section \ref{train_time_section}) indicated that training from random initialization always benefited from increased training time, whereas fine-tuning did not. Additionally, pilot experiments indicated that training from random initialization, but not fine-tuning, benefited from a reduced batch size with very small datasets.

\FloatBarrier
\clearpage
\section{Logistic regression performance of public checkpoints}
\label{public_logreg}

\begin{figure}[h]
    \centering
    \includegraphics[width=\linewidth]{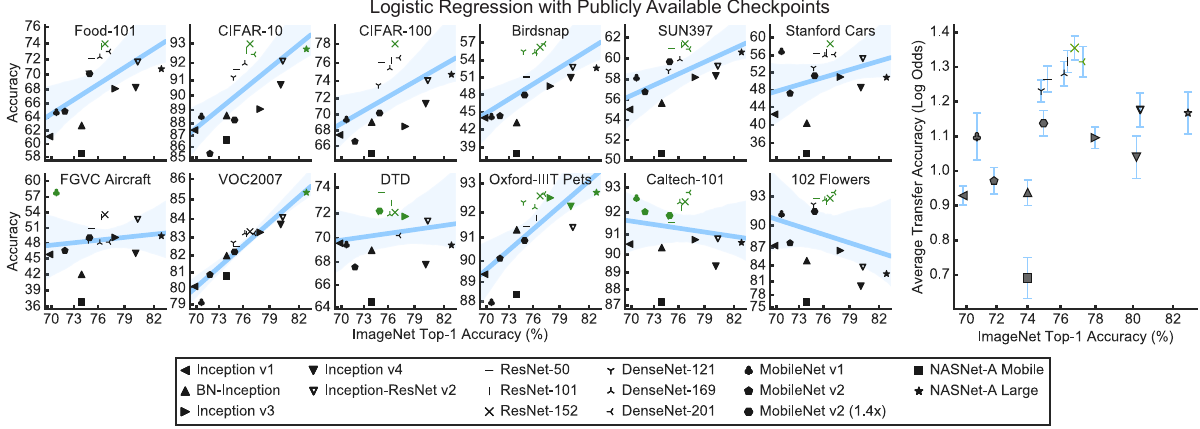}
    \caption{Accuracy of logistic regression classifiers on fixed features from publicly available checkpoints, rather than retrained models. See also Figure \ref{scatterplots}.}
    \label{feature_vector_public_checkpoints}
\end{figure}

We present results of logistic regression with features extracted from publicly available checkpoints in Figure \ref{feature_vector_public_checkpoints}. With these checkpoints, ResNets and DenseNets were consistently among the top performing models. The correlation between ImageNet top-1 accuracy and accuracy across transfer tasks was weak and did not reach statistical significance ($r = 0.37$, $p = 0.16$). By contrast, the correlation with between ImageNet top-1 accuracy and accuracy across transfer tasks with retrained models ($r = 0.99$) was much higher ($p < 10^{-4}$, $z = 5.2$, test of equality of nonoverlapping correlations based on dependent groups \cite{silver2004testing,diedenhofen2015cocor}).

\begin{figure}[h]
    \centering
    \includegraphics{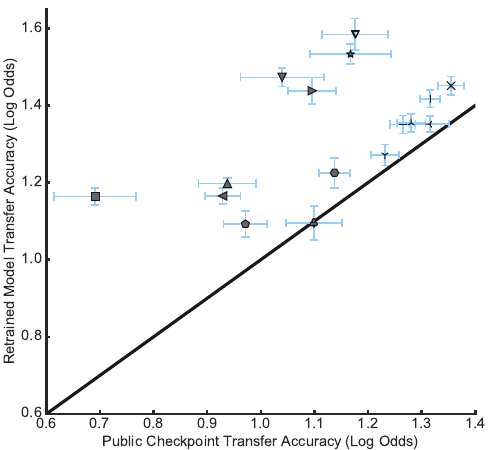}
    \caption{Our retrained models consistently outperform public checkpoints for logistic regression. See Figure~\ref{feature_vector_public_checkpoints} for legend.}
    \label{feature_vector_public_vs_retrained}
\end{figure}

Retrained models achieved higher transfer accuracy than publicly available checkpoints. For 11 of the 12 datasets investigated (all but Oxford Pets), features from the best retrained model achieved higher accuracy than features from the best publicly available checkpoint. Retrained models achieved higher accuracy for 84\% of dataset/model pairs (162/192), and transfer accuracy averaged across datasets was higher for retrained models for all networks except MobileNet v1 (Figure \ref{feature_vector_public_vs_retrained}). The best retrained model, Inception-ResNet v2, achieved an average log odds of 1.58, whereas the best public checkpoint, ResNet v1 152, achieved an average log odds of 1.35 ($t(11) = 5.6$, $p = 0.0002$).

\FloatBarrier
\clearpage
\section{Extended analysis of effect of training/regularization settings}

\subsection{Performance of penultimate layer features}
\label{checkpoints_appendix}

\begin{figure}[h]
    \centering
    \includegraphics{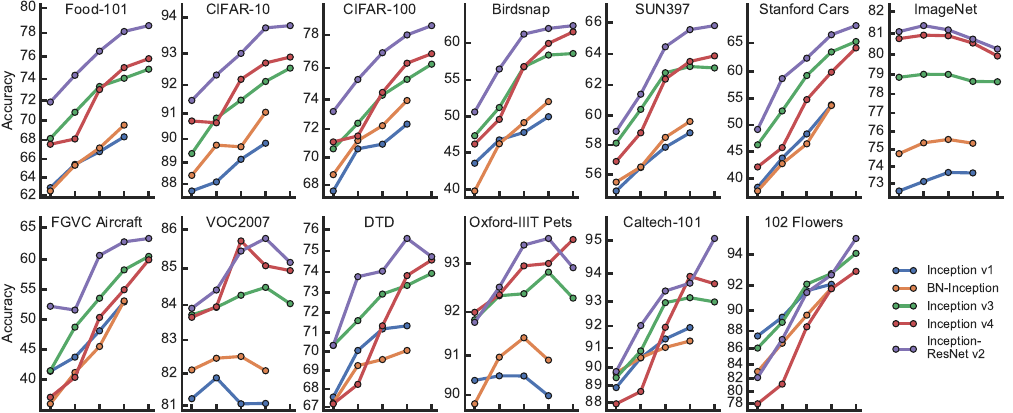}
    \caption{When performing logistic regression on penultimate layer features, all datasets and models benefit from removal of regularization. Each subplot represents transfer performance on one of the datasets investigated. The top right plot shows ImageNet top-1 accuracy of the models. Points along the x-axis represent different training settings (presence/absence of batch normalization scale parameter, label smoothing, dropout, and presence/absence of auxiliary head), following the same convention as in Figure~\ref{model_variants}. The leftmost setting is the Inception default, and uses no batch normalization scale parameter, but includes label smoothing, dropout, and an auxiliary head. From left to right, we enable the batch normalization scale parameter; disable label smoothing; disable dropout; and disable the auxiliary head.}
    \label{model_variants_all_datasets}
\end{figure}

Figure~\ref{model_variants_all_datasets} shows performance of penultimate layer features in each of the training settings in Figure~\ref{model_variants}, broken down by dataset. Across nearly all datasets and models, the least-regularized models achieved the highest performance, even though these models were not the best in terms of ImageNet top-1 accuracy. We also experimented with removing weight decay, but found that this yielded substantially lower performance on both ImageNet and all transfer datasets except for FGVC Aircraft.

\begin{figure}[h]
    \centering
    \includegraphics{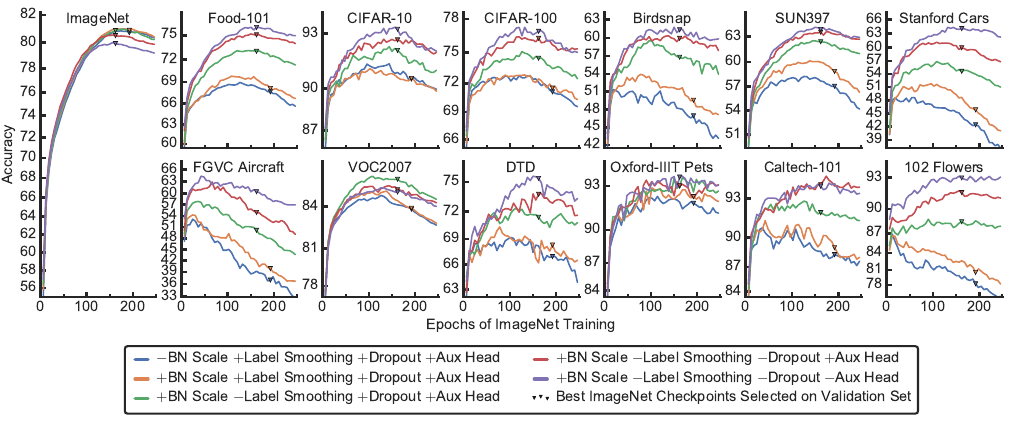}
    \caption{Regularization affects transfer learning with fixed features earlier in training than ImageNet top-1 accuracy. Transfer learning performance of Inception v4 checkpoints evaluated every 12 epochs. Triangles represent the checkpoint that optimized performance on a validation set split from the training set, in a separate training run.}
    \label{model_variants_over_training}
\end{figure}

To investigate whether the effect of regularization upon the performance of fixed features was mediated by training time, rather than the regularization itself, we performed logistic regression on penultimate layer features from Inception v4 at different epochs over training (Figure~\ref{model_variants_over_training}). For models with more regularizers, checkpoints from earlier in training typically performed better than the checkpoint that achieved the best accuracy on ImageNet. However, on most datasets, the best checkpoint without regularization outperformed all checkpoints with regularization. For most datasets, the best transfer accuracy was achieved at around the same number of training epochs as the best ImageNet top-1 accuracy, but on FGVC Aircraft, we found that a checkpoint early in training yielded much higher accuracy.

\subsection{Fine-tuning performance}
\label{finetune_reg_appendix}

\begin{figure}[h]
    \centering
    \includegraphics{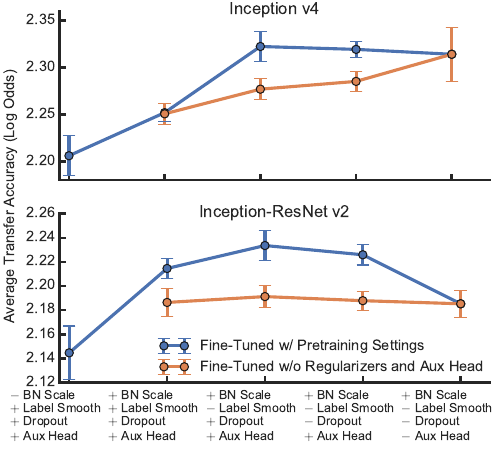}
    \caption{Using regularizers at ImageNet pretraining time does not benefit fine-tuning performance unless the same regularizers are used to fine-tune. Blue points represent models pretrained and fine-tuned with the same training configuration, as in Figure~\ref{ft_settings}. Orange points represent models pretrained with different configurations but fine-tuned without regularization or an auxiliary head (the rightmost configuration in the plot). Accuracy is averaged across 3 fine-tuning runs each for 2 rounds of hyperparameter tuning.}
    \label{ft_settings_cmp}
\end{figure}

\begin{figure}[h]
    \centering
    \includegraphics{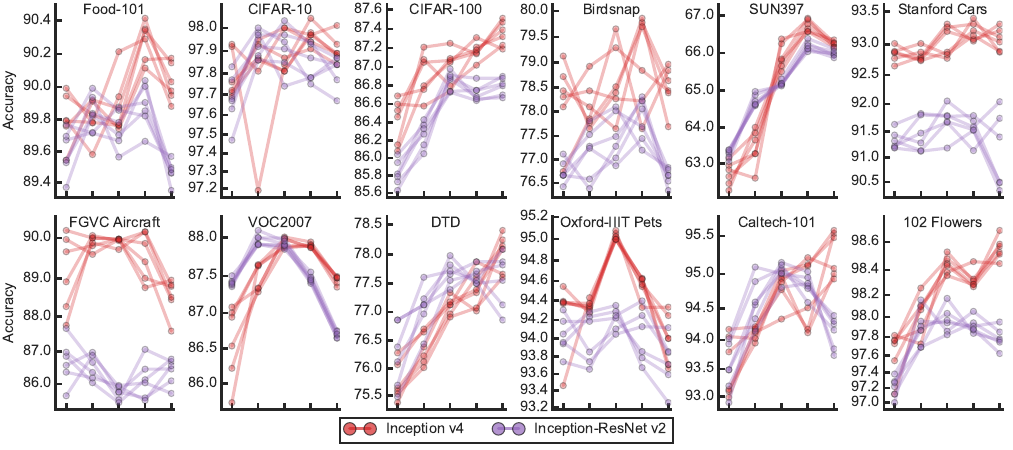}
    \caption{For fine-tuning, different training settings are best on different datasets. Points along the x-axis represent different training settings (presence/absence of batch normalization scale parameter, label smoothing, dropout, and presence/absence of auxiliary head), following the same convention as in Figure~\ref{ft_settings_cmp}. The leftmost setting is the Inception default, and uses no batch normalization scale parameter, but includes label smoothing, dropout, and an auxiliary head. From left to right, we enable the batch normalization scale parameter; disable label smoothing; disable dropout; and disable the auxiliary head. Each line shows performance for a different fine-tuning run.}
    \label{ft_comparison_all_datasets}
\end{figure}

In this section, we present an expanded analysis of the effect of regularization upon fine-tuning analysis. Figure~\ref{ft_settings_cmp} shows average fine-tuning both performance across datasets when fine-tuning with the same settings as used for pretraining (blue, same data as in Figure~\ref{ft_settings}), and when pretraining with regularization but fine-tuning without any regularization (orange). For all regularization settings, benefits are only clearly observed when the regularization is used for both pretraining and fine-tuning. Figure~\ref{ft_comparison_all_datasets} shows results broken down by dataset for pretraining and fine-tuning with the same settings. 

Overall, the effect of regularization upon fine-tuning performance was much smaller than the effect upon the performance of logistic regression on penultimate layer features. As in the logistic regression setting, enabling batch normalization scale parameters and disabling label smoothing improved performance. Effects of dropout and the auxiliary head were not entirely consistent across models and datasets (Figure~\ref{ft_comparison_all_datasets}). Inception-ResNet v2 clearly performed better when the auxiliary head was present. For Inception v4, the auxiliary head improved performance on some datasets (Food-101, FGVC Aircraft, VOC2007, and Oxford Pets) but worsened performance on others (CIFAR-100, DTD, Oxford 102 Flowers). However, because improvements were only observed when the auxiliary head was used both for pretraining and fine-tuning, it is unclear whether the auxiliary head leads to better weights or representations. It may instead improve fine-tuning performance by acting as a regularizer at fine-tuning time.

\FloatBarrier
\section{Relationship between dataset size and predictive power of ImageNet accuracy}

\label{second_order_appendix}

\begin{figure}[h]
    \centering
    \includegraphics[width=\linewidth]{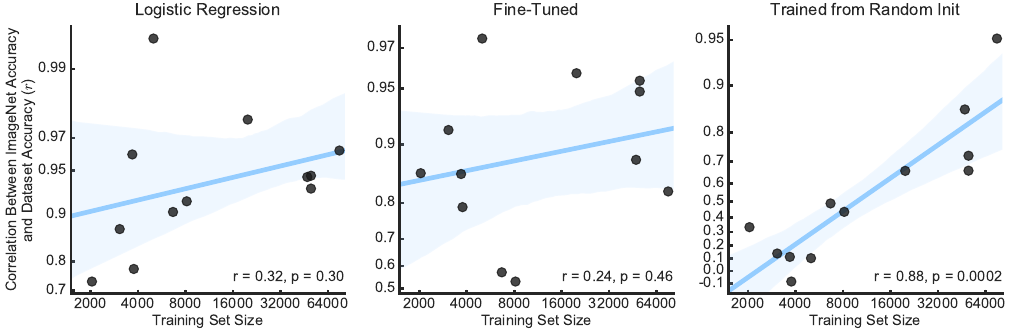}
    \caption{When training from random initialization, the correlation between ImageNet top-1 accuracy and transfer accuracy is higher for larger datasets. Each point represents one of the 12 datasets investigated. The y-axis represents the Pearson correlation between ImageNet top-1 accuracy and accuracy for that dataset, based on the 16 ImageNet networks investigated, and is scaled according to the Fisher z-transformation. The x-axis is log-scaled. $r$ and $p$ values in bottom left reflect the correlation between the log-transformed training set size and Fisher z-transformed correlations.}
    \label{dataset_size_vs_cor}
\end{figure}

As datasets get larger, ImageNet accuracy becomes a better predictor of the performance of models trained from scratch. Figure~\ref{dataset_size_vs_cor} shows the relationship between the dataset size and the correlation between ImageNet accuracy and accuracy on other datasets, for each of the 12 datasets investigated. We found that there was a significant relationship when networks were trained from random initialization ($p = 0.0002$), but there were no significant relationships in the transfer learning settings.

One possible explanation for this behavior is that ImageNet performance measures both inductive bias and capacity. When training from scratch on smaller datasets, inductive bias may be more important than capacity.

\section{Additional comparisons of logistic regression, fine-tuning, and training from random initialization}

\label{comparison_scatters_appendix}

\begin{figure*}[htbp]
    \centering
    \includegraphics[width=\linewidth]{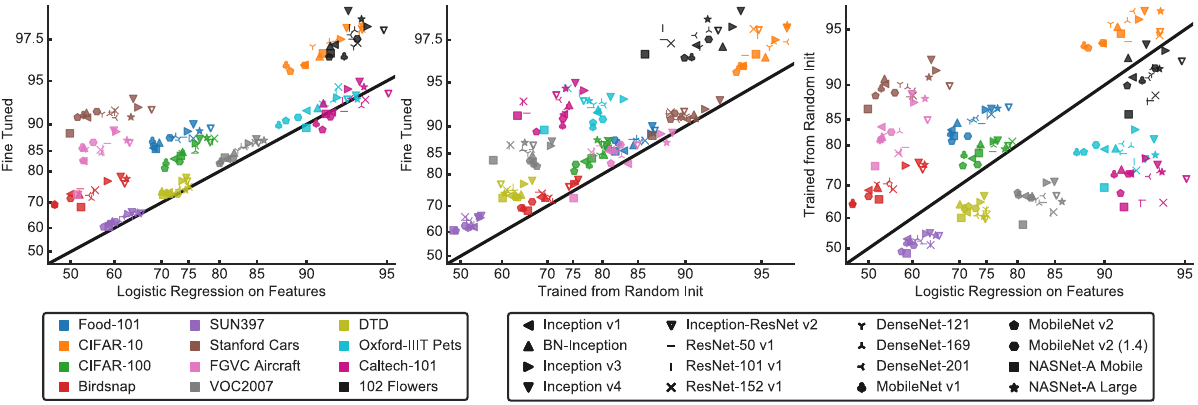}
    \caption{Scatter plots comparing trained models in each pair of settings investigated. Axes are logit-scaled.}
    \label{comparison_scatters}
\end{figure*}

Figure~\ref{comparison_scatters} presents additional scatter plots comparing performance in the three settings we investigated. Fine-tuning usually achieved higher accuracy than logistic regression on top of fixed ImageNet features or training from randomly initialized models, but for some datasets, the gap was small. The performance of logistic regression on fixed ImageNet features vs. networks trained from random initialization was heavily dependent on the dataset.

\FloatBarrier
\section{Comparison versus state-of-the-art}
\label{best_models_appendix}

\begin{table*}[ht]
    \begin{minipage}{\linewidth}
    \small
    \centering
    \begin{tabular}{|l|l|l|l|l|}
    \hline
    & \multicolumn{2}{c|}{Previously reported} & \multicolumn{2}{c|}{Current work}\\
    \multicolumn{1}{|c|}{Dataset}     & \multicolumn{1}{c|}{Acc.} & \multicolumn{1}{c|}{Method} & \multicolumn{1}{c|}{Acc.} & \multicolumn{1}{c|}{Best network}\\
    \hline\hline
    Food-101 & 90.4 & Domain-specific transfer @ 448 \cite{cui2018fgvc} & 90.0 (\textbf{90.8}\textsuperscript{\emph{a}}) & Inception v4, fine-tuned\\
    CIFAR-10 & 98.5 & AutoAugment \cite{cubuk2018autoaugment} & 98.0\textsuperscript{\emph{b}} & NASNet-A Large, fine-tuned\\
    CIFAR-100 & 89.3 & AutoAugment \cite{cubuk2018autoaugment} & 87.5\textsuperscript{\emph{b}} & NASNet-A Large, fine-tuned\\
    Birdsnap & 80.2\textsuperscript{\emph{c}} & Mask-CNN @ 448 \cite{WEI2018704} & 78.4 (\textbf{81.8}\textsuperscript{\emph{a}}) & Inception v4, fine-tuned\\
    SUN397 & \textbf{70.2} & \multicolumn{1}{p{5cm}|}{Places/ImageNet-pretrained multi-scale VGG ensemble \cite{herranz2016scene}} & 66.4 (68.3\textsuperscript{\emph{a}}) & Inception v4, fine-tuned\\
    Stanford Cars & \textbf{94.8} & AutoAugment @ 448 \cite{cubuk2018autoaugment} & 93.3 (93.4\textsuperscript{\emph{a}}) & Inception v4, fine-tuned\\
    FGVC Aircraft & \textbf{92.9}\textsuperscript{\emph{c}} & Deep layer aggregation @ 448 \cite{deeplayeraggregation} & 89.0 (90.9\textsuperscript{\emph{a}}) & Inception v4, fine-tuned\\
    VOC 2007 Cls. & \textbf{89.7} & VGG multi-scale ensemble \cite{simonyan2014very} & 87.4 (88.2\textsuperscript{\emph{a}}) & Inception v4, fine-tuned\\
    DTD & 75.5 & FC+FV-CNN+D-SIFT \cite{cimpoi2015deep} & \textbf{78.1} & Inception v4, fine-tuned\\
    Oxford-IIIT Pets & 93.8 & Object-part attention \cite{peng2018object} & \textbf{94.5} & ResNet-152 v1, fine-tuned\\
    Caltech-101 & 93.4 & Spatial pyramid pooling \cite{he2014spatial} & \textbf{95.1} & Inception-ResNet v2, log. regression\\
    Oxford 102 Flowers & 97.7 & Domain-specific transfer \cite{cui2018fgvc} & \textbf{98.5} & Inception v4, fine-tuned\\
    \hline
    \end{tabular}
{
    \footnotetext[1]{For datasets where the best published result evaluated at $448 \times 448$ or at multiple scales, we provide results at $448 \times 448$ in parentheses.}
    \footnotetext[2]{Accuracy excludes images duplicated between the ImageNet training set and CIFAR test sets; see Appendix \ref{dupimages}. A previous version of this paper achieved accuracies of 98.4\% on CIFAR-10 and 88.2\% on CIFAR-100 by fine-tuning the public NASNet checkpoint with the auxiliary head, dropout, and drop path. The difference in this version is due to the change in settings; the previous results remain valid.}
    \footnotetext[3]{Krause et al. \cite{krause2016} achieve 85.4\% on Birdsnap and 95.9\% on Aircraft using bird and aircraft images collected from Google image search.}}
    \end{minipage}
    \caption{Performance of best models.}
    \label{tab:best_models}
\end{table*}

Table \ref{tab:best_models} shows the best previously reported results of which we are aware on each of the datasets investigated. We achieve state-of-the-art performance on either 4 datasets at networks' native image sizes, or 6 if we retrain networks at $448 \times 448$, as some previous transfer learning works have done. For CIFAR-10, CIFAR-100, and Stanford Cars, the best result was trained from scratch; for all other datasets, the baselines use some form of ImageNet pretraining.

\FloatBarrier
\section{Comparison of alternative classifiers}
\label{svm}
\begin{figure}[h]
    \centering
    \includegraphics{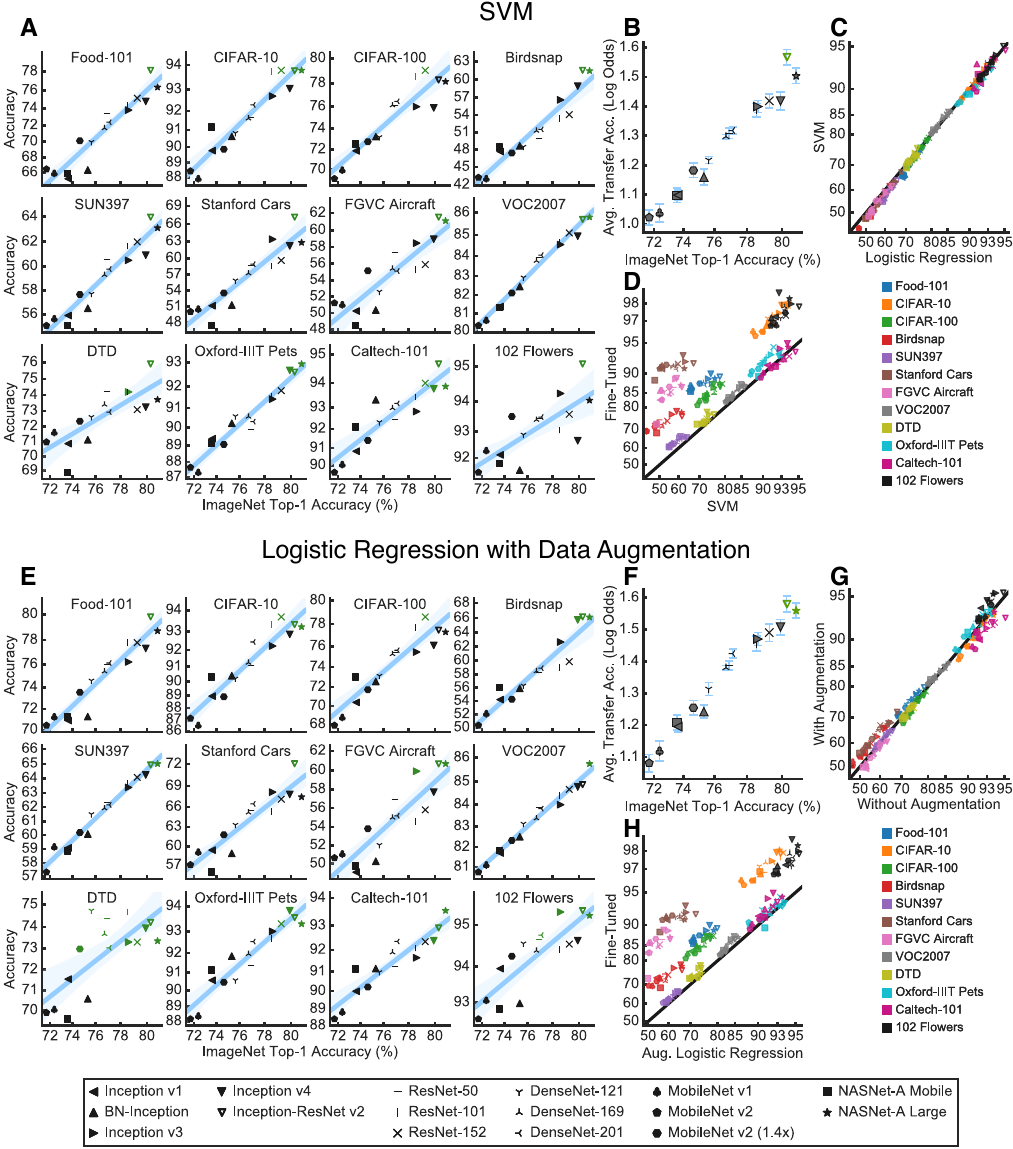}
    \caption{Analysis of SVM and logistic regression with data augmentation, performed on fixed features. \textbf{A} and \textbf{E}: Scatter plots of ImageNet top-1 accuracy versus transfer accuracy on each of the 12 datasets examined. See also Figure~\ref{scatterplots}. \textbf{B} and \textbf{F}: ImageNet top-1 accuracy versus average transfer accuracy for each network investigated. \textbf{C} and \textbf{G}: Performance of logistic regression without data augmentation versus SVM (\textbf{C}) or logistic regression with data augmentation (\textbf{G}). \textbf{D} and \textbf{H}: Performance of SVM (\textbf{D}) or logistic regression with data augmentation (\textbf{H}) versus fine-tuning. See also Figure~\ref{comparison_scatters}.}
    \label{svmres}
\end{figure}
In addition to the logistic regression without data augmentation setting described in the main text, we investigated transfer learning performance using support vector machines without data augmentation, and using logistic regression with data augmentation. Results are shown in Figure~\ref{svmres}.

\subsection{SVM}

Although a logistic regression classifier trained on the penultimate layer activations has a natural interpretation as retraining the last layer of the neural network, many previous studies have reported results with support vector machines \cite{donahue2014decaf,razavian2014cnn,Chatfield14,simonyan2014very}. Thus, we examine performance in this setting as well (Figure \ref{svmres}A-D). Following previous work \cite{simonyan2014very,Chatfield14}, we $L_2$-normalized the input to the model along the feature dimension. We used the SVM implementation from scikit-learn \cite{fan2008liblinear,pedregosa2011scikit}, selecting the value of the hyperparameter $C$ from 26 logarithmically spaced values between $0.001$ and $100$. SVM and logistic regression results were highly correlated ($r = 0.998$). For most (146/192) dataset/model pairs, logistic regression outperformed SVM, but differences were small (average log odds 1.32 vs. 1.28, $p < 10^{-19}$, t-test).

\subsection{Logistic regression with data augmentation}

Finally, we trained a logistic regression classifier with data augmentation, in the same setting we use for fine-tuning. We trained for 40,000 steps with Nesterov momentum and a batch size of 256. Because the optimization problem is convex, we did not optimize over learning rate, but instead fixed the initial learning rate at 0.1 and used a cosine decay schedule. We optimized over L2 regularization parameters for the final layer, applied to the mean of the per-example losses, selected from a range of 10 logarthmically spaced values between $10^{-10}$ and 0.1. Results are shown in Figure~\ref{svmres}E-H. No findings changed. Transfer accuracy with data augmentation was highly correlated with ImageNet accuracy (Figure~\ref{svmres}F) and with results without data augmentation (Figure~\ref{svmres}G; $r = 0.99$ for both correlations). Fine-tuning remained clearly superior to logistic regression with data augmentation, achieving better results for 188/192 dataset/model pairs (Figure~\ref{svmres}H).

Logistic regression with data augmentation performed better for 100/192 dataset/model pairs. Data augmentation gave a slight improvement in average log odds (1.35 vs. 1.32), but the best performing model without data augmentation was better than the best performing model with data augmentation on half of the 12 datasets.

\FloatBarrier
\section{Duplicate images}
\label{dupimages}

\begin{table*}[htbp]
\centering
{
\begin{tabular}{|l|r|r|r|r|r|r|}
\hline
Dataset &  Train Size &  Test Size &  Train Dups &  Test Dups &  Train Dup \% &  Test Dup \% \\
\hline\hline
Food-101         &     75,750 &    25,250 &           2 &          1 &          0.00\% &         0.00\% \\
CIFAR-10         &     50,000 &    10,000 &         703 &        137 &          1.41\% &         1.37\% \\
CIFAR-100        &     50,000 &    10,000 &       1,134 &        229 &          2.27\% &         2.29\% \\
Birdsnap         &     47,386 &     2,443 &         431 &         23 &          0.91\% &         0.94\% \\
SUN397           &     19,850 &    19,850 &         113 &         95 &          0.57\% &         0.48\% \\
Stanford Cars    &      8,144 &     8,041 &          10 &         14 &          0.12\% &         0.17\% \\
FGVC Aircraft    &      6,667 &     3,333 &           0 &          1 &          0.00\% &         0.03\% \\
VOC2007          &      5,011 &     4,952 &          46 &         38 &          0.92\% &         0.77\% \\
DTD              &      3,760 &     1,880 &          14 &          9 &          0.37\% &         0.48\% \\
Oxford-IIIT Pets &      3,680 &     3,669 &         227 &         58 &          6.17\% &         1.58\% \\
Caltech-101      &      3,060 &     6,084 &          28 &         21 &          0.92\% &         0.35\% \\
102 Flowers      &      2,040 &     6,149 &           1 &          0 &          0.05\% &         0.00\% \\
\hline
\end{tabular}
}
\caption{Prevalence of images duplicated between the ImageNet training set and datasets investigated for transfer.}
\label{duptable}
\end{table*}

We used a CNN-based duplicate detector trained on synthesized image triplets to detect images that were present in both the ImageNet training set and the datasets we examine. Because the duplicate detector is optimized for speed, it is imperfect. We used a threshold that was conservative based on manual examination, i.e., it resulted in some false positives but very few false negatives. Thus, the results below represent a worst-case scenario for overlap in the datasets examined. Generally, there are relatively few duplicates. For most of these datasets, standard practice is to fine-tune an ImageNet pretrained network without special handling of duplicates, so the presence of duplicates does not affect the comparability of our results to previous work. However, for CIFAR-10 and CIFAR-100, we compare against networks trained from scratch and there are a substantial number of duplicates, so we exclude duplicates from the test set.

On CIFAR-10, we achieve an accuracy of 98.04\% when fine-tuning NASNet Large (the best model) on the full test set. We also achieve an accuracy of 98.02\% on the 9,863 example test set that is disjoint with the ImageNet training set. We achieve an accuracy of 99.27\% on the 137 duplicates. On CIFAR-100, we achieve an accuracy of 87.7\% on the full test set. We achieve an accuracy of 87.5\% on the 9,771 example test set that is disjoint from the ImageNet training set, and an accuracy of 95.63\% on the 229 duplicates.

\FloatBarrier
\section{Numerical performance results}
\label{model_performance_appendix}

We present the numerical results for logistic regression, fine-tuning, and training from random initialization in Table \ref{model_performance}. Bold-faced numbers represent best models, or models insignificantly different from the best, in each training setting.

\begin{table*}[p]
\centering
\footnotesize
\setlength{\tabcolsep}{4pt}

{\small Logistic regression}
\vspace{3pt}

\begin{tabular}{|l|r|r|r|r|r|r|r|r|r|r|r|r|}
\hline
Network & \multicolumn{1}{l|}{Food} & \multicolumn{1}{l|}{CIFAR10} & \multicolumn{1}{l|}{CIFAR100} & \multicolumn{1}{l|}{Birdsnap} & \multicolumn{1}{l|}{SUN397} & \multicolumn{1}{l|}{Cars} & \multicolumn{1}{l|}{Aircraft} & \multicolumn{1}{l|}{VOC2007} & \multicolumn{1}{l|}{DTD} & \multicolumn{1}{l|}{Pets} & \multicolumn{1}{l|}{Caltech101} & \multicolumn{1}{l|}{Flowers}\\
\hline\hline
Inception v1 & 68.3 & 89.8 & 72.3 & 49.9 & 58.8 & 53.8 & 53.0 & 81.1 & 71.3 & 90.0 & 91.93 & 92.1\\
BN-Inception & 69.5 & 91.0 & 73.9 & 52.0 & 59.6 & 53.7 & 53.1 & 82.1 & 70.1 & 90.9 & 91.32 & 91.8\\
Inception v3 & 74.9 & 92.5 & 76.2 & 58.6 & 63.1 & 65.3 & 60.5 & 84.0 & \textbf{73.9} & 92.3 & 92.98 & 94.1\\
Inception v4 & 75.8 & 92.9 & 76.9 & \textbf{61.4} & 63.9 & 64.2 & 59.9 & 84.9 & \textbf{74.6} & \textbf{93.4} & 93.65 & 93.0\\
Inception-ResNet v2 & \textbf{78.6} & \textbf{93.8} & \textbf{78.5} & \textbf{62.3} & \textbf{65.8} & \textbf{67.9} & \textbf{63.3} & 85.2 & \textbf{74.7} & \textbf{92.9} & \textbf{95.06} & \textbf{94.9}\\
ResNet-50 v1 & 74.1 & 91.8 & 76.0 & 52.2 & 62.5 & 59.5 & 58.5 & 83.5 & \textbf{74.9} & 91.5 & 92.74 & 93.2\\
ResNet-101 v1 & 75.1 & \textbf{93.6} & \textbf{78.9} & 55.3 & 64.0 & 60.1 & 57.4 & 84.5 & \textbf{74.9} & 92.2 & 92.65 & 93.1\\
ResNet-152 v1 & 75.8 & \textbf{93.8} & \textbf{79.2} & 55.7 & 64.1 & 60.2 & 56.9 & 84.8 & \textbf{75.0} & 92.4 & 93.96 & 93.5\\
DenseNet-121 & 72.0 & 90.5 & 73.8 & 51.9 & 60.7 & 57.3 & 53.5 & 82.6 & \textbf{74.8} & 91.2 & 92.13 & 93.3\\
DenseNet-169 & 72.7 & 91.8 & 76.2 & 54.9 & 61.2 & 59.0 & 57.2 & 83.0 & \textbf{73.4} & 92.0 & 93.75 & 93.4\\
DenseNet-201 & 73.2 & 92.2 & 76.4 & 54.2 & 61.9 & 60.3 & 57.1 & 83.6 & 73.2 & 91.4 & 93.15 & 93.1\\
MobileNet v1 & 68.2 & 88.2 & 70.9 & 46.3 & 58.8 & 52.9 & 52.6 & 80.2 & 71.0 & 87.4 & 90.77 & 92.7\\
MobileNet v2 & 68.4 & 88.6 & 70.6 & 46.3 & 57.6 & 51.6 & 52.9 & 80.0 & 71.7 & 88.1 & 91.26 & 91.7\\
MobileNet v2 (1.4) & 71.6 & 89.8 & 73.4 & 50.0 & 60.3 & 56.1 & 55.2 & 81.9 & 73.0 & 89.3 & 91.83 & 93.5\\
NASNet-A Mobile & 68.9 & 91.3 & 73.6 & 52.4 & 58.8 & 49.8 & 51.5 & 80.8 & 70.3 & 90.0 & 91.52 & 91.8\\
NASNet-A Large & 76.9 & \textbf{93.8} & 78.0 & \textbf{62.8} & 65.1 & 63.7 & \textbf{62.8} & \textbf{85.8} & \textbf{74.5} & \textbf{93.5} & 93.89 & 93.8\\
\hline
\end{tabular}

\vspace{6pt}
{\small Fine-tuned}
\vspace{3pt}

\begin{tabular}{|l|r|r|r|r|r|r|r|r|r|r|r|r|}
\hline
Network & \multicolumn{1}{l|}{Food} & \multicolumn{1}{l|}{CIFAR10} & \multicolumn{1}{l|}{CIFAR100} & \multicolumn{1}{l|}{Birdsnap} & \multicolumn{1}{l|}{SUN397} & \multicolumn{1}{l|}{Cars} & \multicolumn{1}{l|}{Aircraft} & \multicolumn{1}{l|}{VOC2007} & \multicolumn{1}{l|}{DTD} & \multicolumn{1}{l|}{Pets} & \multicolumn{1}{l|}{Caltech101} & \multicolumn{1}{l|}{Flowers}\\
\hline\hline
Inception v1 & 85.6 & 96.17 & 83.2 & 73.0 & 62.0 & 91.0 & 82.7 & 83.2 & 73.6 & 91.9 & 91.7 & 97.26\\
BN-Inception & 86.8 & 96.67 & 84.8 & 72.9 & 62.8 & 91.7 & 85.8 & 84.6 & 73.9 & 92.3 & 92.8 & 97.2\\
Inception v3 & 88.8 & 97.5 & 86.6 & \textbf{77.2} & 65.7 & 92.3 & \textbf{88.8} & 86.6 & \textbf{77.2} & 93.5 & \textbf{94.3} & 97.98\\
Inception v4 & \textbf{90.0} & \textbf{97.93} & \textbf{87.5} & \textbf{78.4} & \textbf{66.4} & \textbf{93.3} & \textbf{89.0} & \textbf{87.4} & \textbf{78.1} & 93.7 & \textbf{94.9} & \textbf{98.45}\\
Inception-ResNet v2 & 89.4 & \textbf{97.87} & 86.8 & 76.3 & \textbf{65.9} & 92.0 & 86.7 & 86.7 & \textbf{77.1} & 93.3 & \textbf{93.9} & 97.85\\
ResNet-50 v1 & 87.8 & 96.77 & 84.5 & 74.7 & 64.7 & 91.7 & 86.6 & 85.7 & 75.2 & 92.5 & 91.8 & 97.51\\
ResNet-101 v1 & 87.6 & 97.68 & 87.0 & 73.8 & 64.8 & 91.7 & 85.6 & 86.6 & 76.2 & \textbf{94.0} & 93.1 & 97.94\\
ResNet-152 v1 & 87.6 & \textbf{97.91} & \textbf{87.6} & 74.3 & \textbf{66.0} & 92.0 & 85.3 & 86.8 & 75.4 & \textbf{94.5} & 93.2 & 97.35\\
DenseNet-121 & 87.7 & 97.18 & 84.8 & 73.2 & 62.3 & 91.5 & 85.4 & 85.1 & 74.9 & 92.9 & 91.9 & 97.18\\
DenseNet-169 & 88.0 & 97.4 & 85.0 & 71.4 & 63.0 & 91.5 & 84.5 & 85.9 & 74.8 & 93.1 & 92.5 & 97.86\\
DenseNet-201 & 87.3 & 97.41 & 86.0 & 72.6 & 64.7 & 91.0 & 84.6 & 85.8 & 74.5 & 92.8 & 93.4 & 97.68\\
MobileNet v1 & 87.1 & 96.15 & 82.3 & 69.2 & 61.7 & 91.4 & 85.8 & 82.6 & 73.4 & 89.9 & 90.1 & 96.67\\
MobileNet v2 & 86.2 & 95.74 & 80.8 & 69.3 & 60.5 & 91.0 & 82.8 & 82.1 & 72.9 & 90.5 & 89.1 & 96.63\\
MobileNet v2 (1.4) & 87.7 & 96.13 & 82.5 & 71.5 & 62.6 & 91.8 & 86.8 & 83.4 & 73.0 & 91.0 & 91.1 & 97.52\\
NASNet-A Mobile & 85.5 & 96.83 & 83.9 & 68.3 & 60.7 & 88.5 & 72.8 & 83.5 & 72.8 & 89.4 & 91.5 & 96.83\\
NASNet-A Large & 88.9 & \textbf{98.04} & \textbf{87.7} & \textbf{77.9} & \textbf{66.2} & 91.1 & 87.2 & 87.2 & 74.3 & 93.3 & \textbf{94.5} & \textbf{98.22}\\
\hline
\end{tabular}

\vspace{6pt}
{\small Trained from random initialization}
\vspace{3pt}

\begin{tabular}{|l|r|r|r|r|r|r|r|r|r|r|r|r|}
\hline
Network & \multicolumn{1}{l|}{Food} & \multicolumn{1}{l|}{CIFAR10} & \multicolumn{1}{l|}{CIFAR100} & \multicolumn{1}{l|}{Birdsnap} & \multicolumn{1}{l|}{SUN397} & \multicolumn{1}{l|}{Cars} & \multicolumn{1}{l|}{Aircraft} & \multicolumn{1}{l|}{VOC2007} & \multicolumn{1}{l|}{DTD} & \multicolumn{1}{l|}{Pets} & \multicolumn{1}{l|}{Caltech101} & \multicolumn{1}{l|}{Flowers}\\
\hline\hline
Inception v1 & 83.1 & 94.03 & 77.0 & 68.6 & 53.1 & 90.1 & 83.7 & 66.9 & 61.9 & 79.1 & 73.3 & 90.9\\
BN-Inception & 84.4 & 95.17 & 80.2 & 69.6 & 52.5 & 90.7 & 81.7 & 66.9 & 64.4 & 79.3 & 74.3 & 92.8\\
Inception v3 & 86.6 & 95.61 & \textbf{80.8} & \textbf{75.3} & \textbf{54.9} & 91.6 & 87.7 & 70.8 & 65.1 & \textbf{83.2} & \textbf{77.0} & \textbf{93.5}\\
Inception v4 & \textbf{86.7} & \textbf{96.05} & \textbf{81.0} & \textbf{75.9} & \textbf{55.0} & \textbf{92.7} & \textbf{88.8} & \textbf{70.9} & \textbf{66.8} & 81.2 & \textbf{75.4} & \textbf{93.9}\\
Inception-ResNet v2 & \textbf{87.0} & 94.85 & 79.9 & 74.2 & 54.2 & 89.9 & 84.9 & 67.0 & 59.6 & 76.9 & 71.8 & 92.5\\
ResNet-50 v1 & 84.3 & 94.17 & 78.6 & 68.2 & 51.5 & 88.5 & 79.6 & 64.9 & 62.3 & 78.1 & 65.6 & 87.9\\
ResNet-101 v1 & 85.6 & 94.81 & 79.9 & 69.5 & 51.5 & 88.2 & 78.6 & 61.6 & 62.6 & 76.2 & 64.6 & 87.8\\
ResNet-152 v1 & 85.9 & 94.61 & \textbf{80.8} & 68.9 & 51.1 & 88.6 & 78.2 & 61.9 & 61.1 & 74.0 & 64.9 & 88.7\\
DenseNet-121 & 84.8 & 95.35 & 79.5 & 70.4 & 52.6 & 90.1 & 82.1 & 65.9 & 62.9 & 78.6 & 73.5 & 91.2\\
DenseNet-169 & 84.8 & 95.53 & 80.0 & 71.1 & 53.2 & 89.7 & 82.8 & 64.6 & 61.3 & 79.9 & 73.8 & 91.9\\
DenseNet-201 & 85.3 & \textbf{96.05} & \textbf{80.8} & 70.4 & 52.4 & 89.3 & 78.4 & 66.9 & 60.7 & 80.3 & 72.4 & 90.8\\
MobileNet v1 & 82.4 & 93.88 & 77.9 & 64.8 & 51.8 & 89.6 & 81.1 & 67.2 & 63.1 & 78.5 & 73.0 & 90.5\\
MobileNet v2 & 80.9 & 93.68 & 75.2 & 64.3 & 48.8 & 88.6 & 81.3 & 67.8 & 63.7 & 78.5 & 67.7 & 89.9\\
MobileNet v2 (1.4) & 81.9 & 94.07 & 75.5 & 66.8 & 51.1 & 89.0 & 82.7 & 66.3 & 63.1 & 80.1 & 73.1 & 91.9\\
NASNet-A Mobile & 81.9 & 94.73 & 78.3 & 65.9 & 48.3 & 86.7 & 75.1 & 57.9 & 60.1 & 69.4 & 63.5 & 85.8\\
NASNet-A Large & \textbf{86.8} & \textbf{96.06} & 79.2 & \textbf{75.5} & 54.3 & 90.9 & \textbf{88.2} & 65.2 & 60.5 & 77.8 & 73.6 & 91.8\\
\hline
\end{tabular}
\caption{Model performance}
\label{model_performance}
\end{table*}